\documentclass[final,journal]{IEEEtran}
\usepackage{amsmath,amsfonts,amssymb}
\usepackage{array}

\usepackage{textcomp}
\usepackage{stfloats}
\usepackage{url}
\usepackage{verbatim}
\usepackage{cite}
\usepackage[pdftex]{graphicx}
\usepackage{mathrsfs}
\usepackage{mathtools} 
\usepackage{color}
\usepackage{bm}
\usepackage{comment}
\usepackage{algpseudocode}
\usepackage{algorithm}
\usepackage{float}
\usepackage{flushend}
\usepackage{booktabs}
\usepackage{pifont}
\usepackage{caption} 
\newcommand{\xmark}{\ding{55}}%

\def\cG{\mathcal G}

\def\cK{\mathcal K}

\def\cU{\mathcal U}

\newcommand{\hide}[1]{}
\newcommand{\raf}[1]{(\ref{#1})}

\newcommand{\cV}{\ensuremath{\mathcal{V}}}
\newcommand{\cE}{\ensuremath{\mathcal{E}} }

\newcommand{\ceil}[1]{\ensuremath{\lceil #1 \rceil}}

\def\footnoterule{\relax%
  \kern-5pt
  \hbox to \columnwidth{\vrule width 0.5\columnwidth height 0.4pt\hfill}
  \kern4.6pt}
\usepackage[hang,flushmargin]{footmisc}

\let\existstemp\exists
\let\foralltemp\forall
\renewcommand*{\exists}{\existstemp\mkern5mu}
\renewcommand*{\forall}{\foralltemp\mkern5mu}

\usepackage[colorlinks = true,
            linkcolor = black,
            urlcolor  = blue,
            citecolor = black,
            anchorcolor = blue]{hyperref}
\begin{document}
\title{Towards Autonomous and Safe Last-mile Deliveries with AI-augmented Self-driving Delivery Robots}

\author{Eyad Shaklab*, Areg Karapetyan*, Arjun Sharma,  Murad Mebrahtu,  Mustofa Basri,  Mohamed Nagy,  Majid Khonji,  and Jorge Dias% <-this % stops a space
\thanks{*Both authors contributed equally to this work.}%
\thanks{This work was supported by the Khalifa University of Science and Technology under Award No. CIRA-2020-286.}
\thanks{E. Shaklab, A. Sharma, M. Mebrahtu, M. Basri, M. Nagy, M. Khonji, and J. Dias are with the EECS Department, Khalifa University, Abu Dhabi, UAE. (e-mails: \{eyad.shaklab, arjun.sharma, murad.mebrahtu, mustofa.basri, mohamed.nagy, majid.khonji, jorge.dias\}@ku.ac.ae)}
\thanks{A. Karapetyan is with the EECS Department, Khalifa University, Abu Dhabi, UAE, and with the Division of Engineering, New York University Abu Dhabi. (e-mail: areg.karapetyan@nyu.edu)}
}

\maketitle

\begin{abstract}
In addition to its crucial impact on customer satisfaction, last-mile delivery (LMD) is notorious for being the most time-consuming and costly stage of the shipping process. Pressing environmental concerns combined with the recent surge of e-commerce sales have sparked renewed interest in automation and electrification of last-mile logistics. To address the hurdles faced by existing robotic couriers, this paper introduces a customer-centric and safety-conscious LMD system for small urban communities based on AI-assisted autonomous delivery robots. The presented framework enables end-to-end automation and optimization of the logistic process while catering for real-world imposed operational uncertainties, clients' preferred time schedules, and safety of pedestrians. To this end, the integrated optimization component is modeled as a robust variant of the Cumulative Capacitated Vehicle Routing Problem with Time Windows, where routes are constructed under uncertain travel times with an objective to minimize the total latency of deliveries (i.e., the overall waiting time of customers, which can negatively affect their satisfaction). We demonstrate the proposed LMD system's utility through real-world trials in a university campus with a single robotic courier. Implementation aspects as well as the findings and practical insights gained from the deployment are discussed in detail. Lastly, we round up the contributions with numerical simulations to investigate the scalability of the developed mathematical formulation with respect to the number of robotic vehicles and customers.

%The results of a preliminary deployment of the system in a university campus are reported and analysed.  

%In order to minimize the total latency of deliveries (i.e., the overall waiting time of clients) and cater for real-world imposed operational uncertainties, we formulate the optimization model as a robust extension of the Cumulative Capacitated Vehicle Routing Problem with Time Windows, referred to as {\sc RCCVRPTW}. 

\end{abstract}


% \fi
\IEEEpeerreviewmaketitle

\section{Introduction}\label{introo}

\IEEEPARstart{T}{he} unparalleled rise of e-commerce activities over the past decade coupled with continually increasing urbanization have notably hiked the volume of urban freight logistics, especially in the last leg of distribution from suppliers to customers, known as the last-mile delivery (LMD). Within the overall chain, LMD holds the key to customer satisfaction, yet is notoriously cumbersome and inefficient due mainly to: (1) lack of automation; (2) inconsistent and congested delivery routes; (3) consumers' elevated expectation for same-day delivery. These setbacks render LMD as the most expensive step in the supply chain, accounting for up to 53\% of the total shipment cost~\cite{world_eco}. In response, substantial efforts, both in academia and industry, have been invested into the search for more sustainable, optimized, and economic parcel delivery concepts and services~\cite{inbook}.

While deployment of unmanned LMD modes was somewhat limited prior to COVID-19 pandemic, practitioners report an unprecedented surge in autonomous delivery services since the virus hit. Case in point, a San Francisco-based firm Starship Technologies in 2021 celebrated its one-millionth autonomous delivery and in 2022 upped the number to over $3$ millions~\cite{str}. Added to that, logistic giants like DHL and UPS, aiming to future-proof their business, have recently begun active operational trials of electric delivery robots and vehicles~\cite{dhl, ups}. In fact, according to a recent Gartner report~\cite{gartner}, the number of robotic LMD couriers operated worldwide is estimated to reach to one million by 2026.

%, suggests that there will be  autonomous LD operated worldwide by 2026 and this is projected to reach a market value of US\$119 billion by 2031 as per Visiongain report in Q3 of 2021 \cite{visiongain}.          

\begin{figure}[!t]
	\centering
	\includegraphics[trim={0.28cm 0cm 2.1cm 0cm},clip, width=\columnwidth]{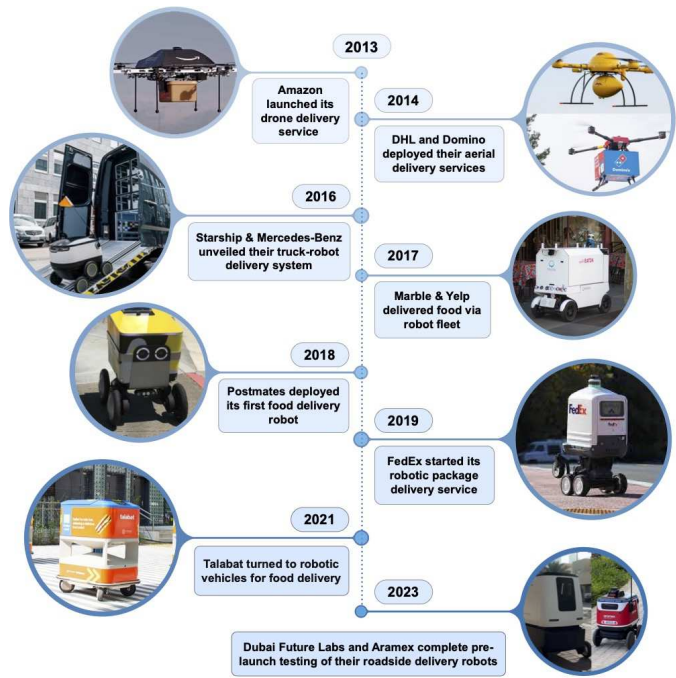}
	\caption{Examples of real-world deployed unmanned LMD services in chronological order since 2013. The observed trend shifts from drone-based approaches to ground vehicles.}
	\label{frm22}
\end{figure}
As summarized in Fig.\ref{frm22}, early attempts of automating parcel deliveries relied primarily on unmanned aerial vehicles (UAVs), commonly known as drones. Admittedly, off-the-shelf drones are agile, swift, low-cost and are being increasingly utilised throughout the globe as a means to streamline monitoring, inspection, mapping, and environmental surveillance procedures~\cite{9779119}. However, their operation in urban areas is accompanied by several challenges and negative byproducts, such as noise pollution, privacy and security concerns, and no-fly zone restrictions~\cite{article_drone_barriers}. As experimentally demonstrated in~\cite{KIRSCHSTEIN2020102209}, drones are energy-efficient specifically for transportation of lightweight packages and could be useful in rural regions, whereas for urban environments ground robots tend to be relatively more energy-saving. Evidently, as deduced from Fig.~\ref{frm22}, the preference for ground-based option has prevailed over time, becoming the mainstream model of LMD robots in the market.

%This is also implied by the tendency of industrial giants like Amazon and others which started shifting towards ground-based solutions as in article \cite{weinberg_2020}.

%While drones could be energy efficient in certain scenarios, they are typically constrained by their meager load-ability and limited battery capacity. 

In developing an automated LMD system with robotic vehicles, the foremost task is to assure the robots can navigate \textit{autonomously} and \textit{safely} in complex dynamic environments. Currently, most of the practically operating units partially rely on remote human control or supervision (e.g., to steer around complex obstacles or halt the navigation to avoid an undetected collision)~\cite{durbin_2021}. Among few exceptions is the food-tech startup Serve Robotics, which in 2022 completed commercial deliveries at Level 4 autonomy (i.e., relying solely on vehicles' onboard capabilities)~\cite{mehar_by_mehar_2022}. It should however be noted that the company's fleet of self-driving LMD carriers is designated to travel specifically on sidewalks.

%For urban spaces, such as sidewalks, 
%Some few exemption however, as cited from the article  The autonomous sidewalk delivery company, Serve Robotics made an achievement for its latest generation of robots which are able to operate dependently without human intervention. These robots can rely on their onboard capabilities to ensure safe operation. 

%However, the current solutions that are practically implemented still require remote operators. They have to keep tabs on the robot fleets at a time even though, arguably these are rare situations where they have to hit brakes or steer around an obstacle as reported in \cite{durbin_2021}. %

%Importantly, a critical factor for autonomous robots is the safety aspect as it is highly prioritized by Starship. An example is the safety mechanism where it could avoid collision with vehicles or hit human. This needs to be considered in while designing the overall system. 

As real-world practice revealed, compared to human counterparts existing robotic LMD couriers fall short in two subtle yet essential aspects -- \textit{flexibility} and \textit{situational inference}. In particular, users often found it inconvenient to attend to the robot for receiving the parcel instead of collecting it from the desired drop-off location~\cite{durbin_2021}. This is also pertinent to current COVID-19 regulations, which endorse contactless operations. More importantly, their lack of basic inference and proactiveness (or interactiveness) with the surrounding environment can expose to otherwise avoidable risks. For instance, in 2019 an incident in Pittsburgh involving an autonomous sidewalk LMD vehicle left a wheelchair-bound individual stranded in a dangerous situation on the street after persistently blocking their passage~\cite{blb}. These demerits, besides limiting the utility of robotic couriers, hinder customers' satisfaction and acceptance of automated last-mile logistic services.

%in automated LMD vehicles
%threat incident reported in

%one of the major inconveniences 
%They’re also inflexible. A customer can’t tell a robot to leave the food outside the door, for example. And some big cities with crowded sidewalks, like New York, Beijing and San Francisco, aren’t welcoming them.

%tarship briefly halted service at the University of Pittsburgh in 2019 after a wheelchair user said a robot blocked her access to a ramp. But the university said deliveries resumed once Starship addressed the issue

%that occurred in the University of Pittsburgh in 2019 during which the  

%Moreover, it transpired from the article in, stating that major aspect users found inconvenient; the delivery which wasn't fully automated and they had to go to enter the pin code to get the parcel, whereas they could be simply unavailable. This is particularly pertinent to current covid regulations, which endorse contact-less operations.

Actuated by the above causes and practical insights, we design and experimentally demonstrate a \textit{fully autonomous AI-boosted} LMD system for optimal and green logistics in urban (small) communities. The system employs self-driving ground vehicles, built on top of a commercial mobile robotic platform, which feature \textit{a touchless} multi-parcel delivery mechanism and a \textit{triple-layer safety stack} (people-platform-package). More concretely, the contributions and roadmap of the present work are as follows:

\begin{enumerate}
	\item After surveying the related literature on robot-assisted LMD approaches in Sec.~\ref{sec:litreview}, we proceed to formalizing the optimization problem involved in the proposed LMD system. Specifically, Sec.~\ref{prob} formulates a robust variant of the Cumulative Capacitated Vehicle Routing Problem with Time Windows, referred to as {\sc RCCVRPTW}, in which a fleet of robotic LMD carriers with varying capacities strive to optimally meet customers' desired delivery schedules in the presence of nondeterministic delays. Here, incorporation of nondeterminism reflects the practical need to account for autonomy-induced uncertainties (e.g., sensor noise, imprecise localization) and external disturbances linked to pedestrian movements or dynamic obstacles.
 
% In {\sc RCCVRPTW}, a fleet of robotic LMD vehicles with given capacities  should satisfy a set of transportation requests while respecting various constraints in relation with the precedence between suppliers and customers, the capacity of vehicles, the opening and closing times of each site.

    % frame the optimal management of automated LMD systems in Sec.~\ref{prob} as an \textit{extended variant} of the Cumulative Capacitated Vehicle Routing Problem with Time Windows. In particular, to incorporate the uncertainities inherent to the operation of autonomous vehicles (e.g., sensor noise, imprecise localization) and external disturbances linked to human-driven vehicles or pedestrians, a robust extension of the problem is devised.

%\footnote{Aleatoric uncertainty refers to unknown factors that can be modeled probabilistically (e.g., sensor noise or motion noise), whereas epistemic uncertainty represents imprecise knowledge }

%epistemic (systematic) and aleatoric (statistical) 
 
 %Optimal logistic scheduling algorithm. To efficiently distribute the packages and reduce the cost of delivering, we need to choose the most desirable delivery plan to execute based on our proposed algorithm. 
	\item Sec.~\ref{sys} provides an in-depth overview of the proposed framework's architecture, followed by the details of the constructed self-driving LMD carrier and its three auxiliary modules for enhanced usability and added safety: (i) A proactive audio service, hinging on a Recurrent Neural Network, for \textit{real-time intent prediction} and early notification of pedestrians; (ii) A reactive perception system for \textit{vibration monitoring} and parcel safety; (iii) Automatic package unloading mechanism. To facilitate reproducibility, the software implementation was based entirely on the open-source Robotic Operating System (ROS). The complete software stack can be accessed online at \url{https://github.com/AV-Lab/AVL_segwayrmp}.

    %Sec.~\ref{sys}, Safety stack (SS). To ensure overall LD safety, we propose the following 3 level safety stack. The upper level is human safety, platform safety comes in the mid-level, and the last level is package safety.
	\item To validate the effectiveness and practicality of the featured autonomous LMD system, a series of test trials were designed and carried out in the outdoor premises of a university campus. The results of preliminary deployment\footnote{Video records of one of the trial runs can be accessed online at \url{https://youtu.be/-9Zams67hNY}.} (with one vehicle), reported in Sec.\ref{sec:exp}, illustrate the merits of the proposed framework allowing the autonomous robotic courier to navigate safely and intelligently in the presence of pedestrians, cars, and dynamic obstacles. In all the case studies performed, the robot completed the autonomous deliveries with minimum total latency without violating customers' preferred time schedules. Moreover, with the newly presented parcel safety module in place, up to $40\%$ reduction of vibrations were recorded in the robot's storage compartment. Lastly, numerical simulations are conducted to analyse the scalability of {\sc RCCVRPTW}'s optimization model against the number of robotic couriers and customers.
 %Contact-less delivery mechanism. To achieve fully autonomous LD we need pragmatic mechanical design. This is more prominent during covid as we're starting to see contact-less proof of delivery becomes standard. 
\end{enumerate}

%seeking to minimize the total latency of deliveries while accounting for customers' preferred time windows. The framework employs self-driving touchless delivery robots 

%Toward achieving fully autonomous, economic, and safe LD, we identify the following 3 criteria.

%Given the above background, this paper develops a prototype LD solution with the ground vehicle and implements it in a real-world environment. Our contributions are including the solution for the 3 criteria mentioned above, proposing the overall concept of the last-mile delivery of fully autonomous vehicles with mechanical design and an algorithm to solve the optimal delivery scheduling problems. The algorithm is designed to have a more realistic scenario by considering the vehicle battery's recharging features as one of its vital constraints. Another contribution is the 3 level safety stacks for that ensures overall safety (human, platform, and package safety). A deep learning-based perception system is used for assessing path smoothness and a modified Generative Adversarial Network (GAN) for pedestrian trajectories prediction, as a pipeline in the system for interactive vehicle-human to ensure human safety.

\begin{table*}[!ht]\label{tab:freq}

\centering
\resizebox{\textwidth}{!}{
\begin{tabular}{|l|l|l|l|l|l|l|l|}
\toprule
                                                & \textbf{System Setup} & \textbf{Objective}                                    & \textbf{Auxiliary Safety Module(s)} & \textbf{Field Testing}     & \textbf{Mathematical Formulation} & \textbf{Delivery Time Constraints} & \textbf{Uncertainty}     \\ \hline
Torabbeigi et al.  \cite{Torabbeigi2020}        & Drones                & Minimize the number of delivery drones                 & {\color{black} $-$}               & {\color{red} \xmark}       & Extended VRP                      & {\color{red} \xmark}               & {\color{red} \xmark}     \\
Dorling et al. \cite{7513397}                   & Drones                & Minimize the overall delivery time or total cost      & {\color{black} $-$}               & {\color{red} \xmark}       & MTVRP                             & {\color{red} \xmark}               & {\color{red} \xmark}     \\
Shao et.al  \cite{9099809}                      & Drones                & Minimize the flight path length of the drone           & {\color{black} $-$}               & {\color{red} \xmark}       & BMCDRP                            & {\color{red} \xmark}               & {\color{red} \xmark}     \\
Huang et al.  \cite{9151388}                    & Bus + Drone           & Minimize delivery time                                & {\color{black} $-$ }              & {\color{red} \xmark}       & VRP                               & {\color{red} \xmark}               & \color{green} \checkmark \\
Mathew et al. \cite{7194856}                    & Truck + Drones        & Minimize the total delivery cost                      & {\color{black} $-$}               & {\color{red} \xmark}       & HDP                               & {\color{red} \xmark}               & {\color{red} \xmark}     \\
Poikonen et al. \cite{Poikonen2017}             & Truck + Drones        & Minimize the delivery time and return of all vehicles & {\color{black} $-$}               & {\color{red} \xmark}       & VRPD                              & {\color{red} \xmark}               & {\color{red} \xmark}     \\
Liu et al. \cite{8989960}                   & Truck + Drones        & Minimize overall travelling cost                      & {\color{black} $-$}               & {\color{red} \xmark}       & E2-VRP                            & {\color{red} \xmark}               & {\color{red} \xmark}     \\
%Ribeiro et al. \cite{9462603} & VRPTW         &Truck + Drones  & Minimize delivery time             & {\color{green} \checkmark}             & {\color{red} \xmark}             &{\color{black} $-$}                &{\color{red} \xmark}                    \\
Boysen et al. \cite{BOYSEN20181085}             & Truck + AGVs          & Minimize weighted number of late deliveries            & {\color{red} \xmark}              & {\color{red} \xmark}       & TBRD                              & Deadline                           & {\color{red} \xmark}     \\
Simoni et al. \cite{SIMONI2020102049}           & Truck + AGV           & Minimize delivery time                                & {\color{red} \xmark}              & {\color{red} \xmark}       & TSP-R                             & {\color{red} \xmark}               & {\color{red} \xmark}     \\
Chen et al. \cite{CHEN20211164}                 & Truck + AGVs          & Minimize total time spent in all routes               & {\color{red} \xmark}              & {\color{red} \xmark}       & VRPTWDR                           & Time Window                        & {\color{red} \xmark}     \\
Chen et al. \cite{CHEN2021102214}               & Truck + AGVs          & Minimize total time spent in all routes               & {\color{red} \xmark}              & {\color{red} \xmark}       & VRPTWDR                           & Time Window                        & {\color{red} \xmark}     \\
Sonneberg et al. \cite{sonneberg2019autonomous} & AGV                   & Minimize delivery cost                                & {\color{red} \xmark}              & {\color{red} \xmark}       & ELRP-TW                           & Time Window                        & {\color{red} \xmark}     \\
Buchegger  et al. \cite{8569339}                & AGV                   & Parcel delivery                                       & {\color{red} \xmark}              & {\color{green} \checkmark} & $-$                               & $-$              & $-$   \\
Li et al. \cite{9238131}                        & AGVs                  & Parcel delivery                                       & \color{green} \checkmark          & {\color{green} \checkmark} & $-$                               & $-$               & $-$      \\
Gao et al. \cite{9901631}                       & AGV                   & Parcel delivery                                       & {\color{red} \xmark}              & {\color{green} \checkmark} & $-$                               & $-$                & $-$      \\ \hline

\hline\hline
Present work                                    & AGVs                  & Minimize customers' waiting time                      & {\color{green} \checkmark}        & {\color{green} \checkmark} & RCCVRPTW                          & Time Window                        & \color{green} \checkmark \\
 \bottomrule
\end{tabular}
}
\caption{A summarised comparison of present work versus prior studies on unmanned LMD systems.}
\label{fig:ComparisonLR}
\end{table*}

\section{Literature Synopsis}\label{sec:litreview}

Given the breadth of literature on last-mile logistics and associated vehicle routing problems, the following survey is confined to studies focusing on unmanned delivery systems. For a more exhaustive review on general LMD problems and approaches, the reader is referred to~\cite{su14095329, doi:10.1080/00207543.2018.1489153, Konstantakopoulos2022}. Depending on the pursued mode of operation, the existing studies on unmanned LMD systems can be organized under three main themes: aerial, hybrid (aerial-ground), and ground-based. In the paragraphs to follow, we survey the developments in each of these research threads, whereas Table \ref{fig:ComparisonLR} provides an overall comparison between the present work and the literature reviewed.

As noted in Sec.~\ref{introo}, early approaches to automating last-mile deliveries focused primarily on drones. For instance, Dorling et al.\cite{7513397} propose a drone-based LMD system and formulate different variants of multi-trip VRPs (MTVRP) that cater for the effect of battery and payload weight on drones' energy consumption while also allowing multiple trips to the depot. The problems are modeled as Mixed-integer linear programs (MILPs) and tackled by the numerical optimizer CPLEX. Focusing on short-distance delivery with drones, Torabbeigi et al.\cite{Torabbeigi2020} model the LMD problem as an extension of the Vehicle Routing Problem (VRP). The problem includes drone battery endurance and drone weight capacity as constraints. On the other hand, Shao et al.\cite{9099809} envision an LMD system with battery exchange stations to extend the drone's delivery coverage range. The optimisation component within the system is defined as the Battery and Maintenance Constrained Drone Routing problem (BMCDRP) where the objectives are to minimize the delivery time and the number of landing depots on the flight path. A heuristic solution methodology based on the Ant Colony and A$^*$ algorithms is developed to find approximate solutions for BMCDRP.

Incidentally, a separate research stream has been exploring hybrid unmanned LMD systems consisting for example of drone-truck, drone-public bus, or drone-AGV combinations. By utilizing the public transportation system (public buses) and drones, Huang et al.\cite{9151388} propose a hybrid LMD system for cost-effective and environmentally friendly long-distance parcel deliveries. The approach employs a Dijkstra-based method to determine the shortest route from the depot to the customer, minimizing the delivery time. Importantly, the proposed approach incorporates probabilistic edge costs to account for the uncertainty in the public transportation network. In \cite{7194856}, Mathew et al. present an LMD system combining a UAV and a truck for deliveries in urban environments. The truck, which is restricted to travel along a street network, transports the drone which delivers packages to customers. The problem is formulated as an optimal path planning problem on a graph, termed as Heterogeneous Delivery Problem (HDP), that falls under the category of a static VRP with coordination constraints between the vehicles. In a similar setup, Poikonen et al. \cite{Poikonen2017} propose an LMD system synergising UAVs and trucks to parallelize deliveries and model the problem as a VRP with drone (VRPD). Liu et al.\cite{8989960} consider an extended variant of the traditional two-echelon VRP (E2-VRP) for cooperated trucks and drones where multiple parcels can be delivered in one drone’s flying route. Subsequently, a two-stage route-based heuristic approach is developed to optimize both the truck’s main route and the drone’s flying routes. 

In support of a fully ground-based realisation, the papers in\cite{BOYSEN20181085, SIMONI2020102049, CHEN20211164, CHEN2021102214} studied various optimization problems for LMD systems comprising trucks/vans loaded with on-board autonomous small robots. Boysen et al.\cite{BOYSEN20181085} propose a variant of VRP, termed as Truck-based Robot Delivery Scheduling problem (TBRD), seeking to minimize the weighted number of late customer deliveries. The studied LMD system considers a delivery truck that launches robotic delivery vehicles each intended for a single customer. The authors develop scheduling procedures that determine the truck route along with the robot depots and drop-off points where robots are launched. In \cite{SIMONI2020102049}, Simoni et al. formulate the LMD problem as an extension of TSP referred to as TSP with Robot (TSP-R). A system consisting of a truck and a robot is envisioned, where the robot is deployed by the truck during delivery stops to serve certain customers. The robot is designed to have a divisible storage compartment and can perform more than one consecutive delivery. In a similar setup, in \cite{CHEN20211164} and \cite{CHEN2021102214}, Chen et al. formulate the LMD problem as a VRP with time windows and delivery robots (VRPTWDR) where the objective is to minimize the total route duration. Time window constraints are incorporated to account for customers' availability. In view of VRPTWDR's NP-hardness, an adaptive large neighborhood search heuristic algorithm is proposed as a solution method.

Towards complete automation of last-mile logistics, several prior works developed and/or studied LMD systems based exclusively on AGVs. In \cite{sonneberg2019autonomous}, Sonneberg et al. frame the LMD problem as a version of VRP coined as Electric Location Routing Problem with Time Windows (ELRP-TW). In this model, customers are assigned to stations, each with its own AGV. The objective in ELRP-TW seeks to optimize the total cost by determining the optimal positioning of stations as well as route construction of AGVs. Moreover, battery, parcel weight, and volume restrictions are considered for multiple consecutive deliveries. On the applied side, Buchegger et al. \cite{8569339} present an autonomous vehicle, built upon a commercially available electric vehicle (Jetflyer), for parcel delivery in urban environments. The vehicle was evaluated in two urban settings. However, the developed system employs a hierarchical first-come-first-serve approach with no optimization of delivery routes and order. In \cite{9901631}, Gao et al. developed a bespoke self-driving robot for last-mile deliveries. The software implementation was built upon Autoware, which is an open-source framework for autonomous driving. In \cite{9238131}, Li et al. present JD.com's (a large e-commerce company in China) methodological framework for autonomous last-mile deliveries in complex traffic conditions and outline the architecture of their autonomous delivery vehicle. The vehicle features a machine learning-based perception module for recognizing generic traffic-related objects/signs and tracking dynamic obstacles in the environment (e.g., vehicles and pedestrians). To navigate safely through complex traffic conditions, the system employs a prediction module to approximate the future trajectories of the tracked moving obstacles as well as a remote monitoring and control module.

As transpires from the above review and Table~\ref{fig:ComparisonLR}, the existing literature studied autonomous LMD systems from one specific angle, focusing either on the underlying mathematical optimization model or the technical and implementation aspects of self-driving robotic vehicles. To close this gap, we unify these two constituents into a coherent LMD system, thereby  enabling end-to-end automation
and optimization of the logistic process. In turn, practical insights gained during the implementation and deployment served to inform the mathematical modeling of the optimization component as reflected by the incorporation of non-determinism to account for autonomy-induced uncertainties and real-world externalities. Furthermore, the current study advances the existing research by introducing two new modules for refined usability and added safety, namely the reactive perception system for vibration monitoring and parcel safety and the automatic package unloading mechanism for contactless delivery.

%Even though such studies conducted experimental validation for the autonomous delivery scenario, they did not implement multi-parcel delivery with the contact-less delivery mechanical design, which allows for fully autonomous delivery without human interventions. Furthermore, the above authors mentioned that the safety aspects were limited to the robot's navigation safety. Our proposed approach implemented a more robust safety aspect to ensure overall safety. 

%CRITICAL ATTACK ON AFOREMENTIONED PAPERS:
%However, none of the above studies conducted experimental validation and did not consider the safety aspect of their approaches.
%Apart from the experimental validation aspect, the problem formulations proposed in the surveyed papers do not account for uncertainties.

%TOTAL PAPER (AS in 27/11): 12 + 2(NOT RELATED) + 1(POSSIBLY REMOVE) 


% \usepackage{booktabs}
% \usepackage{booktabs}
% \usepackage{longtable}
% Note: It may be necessary to compile the document several times to get a multi-page table to line up properly

\section{Problem Statement}\label{prob}
\begin{figure}[!t]
	\centering
	\includegraphics[trim={0cm 0.21cm 0cm 0cm},clip, width=\columnwidth]{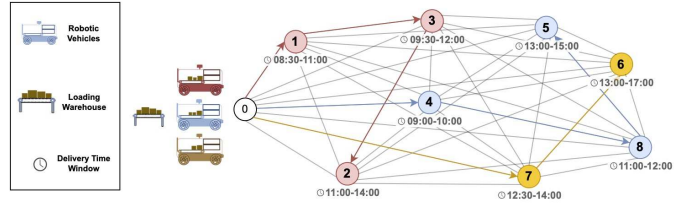}
	\caption{An abstract illustration of the studied LMD system where a fleet of robotic couriers serve a set of customers geographically dispersed around the central depot (node 0).}
	\label{frm1}
\end{figure}

Recall that within the proposed architecture, sketched in Fig.~\ref{frm1}, {\sc RCCVRPTW} serves as a logistic planner responsible for optimizing the dispatch of contracted delivery robots such that the waiting time of all customers is minimized. The robotic couriers must depart from the central depot and visit the respective set of customers\footnote{Each customer is visited once and the delivery service can be performed only within specific customer-defined time windows.} assigned by {\sc RCCVRPTW} such that collectively all the delivery requests are fulfilled. To define the problem formally, we next establish the relevant mathematical notation.

 %which minimizes the sum of the elapsed times (or latencies) to reach a given set of nodes.
 
  %Vehicle Routing Problem with Time Windows, referred to as {\sc RCCVRPTW}.
%poses the challenge of finding
%the optimal route of a vehicle delivering products to multiple locations. 

%used to design an
%optimal route for a fleet of vehicles to service a set of customers, given a set of constraints

%What is the optimal set of routes for a fleet of drones
%to serve a given set of customers

%ow to serve a set of customers, geographically dispersed around the central depot, using a fleet of trucks with varying capacities. 

%optimizing the route cost of a vehicle delivering products to several locations.

%In essence, the problem involves the satisfaction of a set of load transportation requests (jobs) by a fleet of vehicles housed at a depot. 

%This section is devoted to mathematical modeling of {\sc RCCVRPTW}.  whereas the AI stack (detailed in Sec.~\ref{sys}) deals with autonomous 'local' navigation and the three safety principles outlined above{\color{red}[R]}.

Consider a directed or undirected complete graph  $\cG = \big(\cV \cup \{0\}, \cE\big)$, where the root $0$ harbours the depot, the vertices in $\cV \triangleq \{1, ..., n\}$ encode customer destinations and the edge set $\cE \triangleq \big\{ (i,j) : i, j \in \cV \cup \{0\}, i \neq j\big\}$ corresponds to the delivery routes between these nodes. Each arc $(i,j) \in \cE$ is associated with a duration $d_{i,j}$ that measures the robotic vehicles' travel time while traversing from site $i$ to $j$. To synthesize a logistic planner that is robust against nondeterminism stemming from unknown parameters or external disruptions (e.g., imprecise localization, sensor noise, number of encounters with pedestrians), we model $d_{i,j}$ as a stochastic, yet bounded, parameter with a box uncertainty set of the form  $\cU = \big\{d_{i,j} \in \mathbb{R}_{+} :  |d_{i,j} - \bar{d}_{i,j}| \leq \epsilon\big\}$, where $\bar{d}_{i,j}$ denotes the \textit{nominal travel time} and $\epsilon \in \mathbb{R}_{+}$ is a preset constant. A user-selected time interval $[s_i, f_i]$ is imposed on every vertex $i \in \cV$, with $s_i$ and $f_i$ specifying the earliest and latest times the delivery service at node $i$ can be performed, respectively (i.e., the availability window of a customer). Should a robotic courier reach vertex $i$ prior to start time $s_i$, a delay occurs before the package is supplied. The \textit{latency} $l_i$ of node $i \in \cV$ is defined as the total elapsed time (since departing from the depot) until the service at node $i$ starts.

\if
Let $l_i$ denote the For each node $i \in \cV$, denote its latency

 define by $l_i$ its \textit{latency}; the total time elapsed since departing from the depot and the beginning of service for customer $i$, denoted by $t_i$ (naturally, $l_0 = 0$). {\sc RCCVRPTW} asks for a tour $\pi$ on $\cV \cup \{0\}$ (i.e., a permutation of these vertices), rooted at node $0$ (i.e., $\pi(0) = 0$) and containing all the nodes in $\cV$, which minimizes 
\begin{eqnarray}
\sum_{i=1}^n l_i & \triangleq &\sum_{i=1}^{n}\sum_{j=1}^{i} (t_{\pi(j)} - t_{\pi(j-1)})  \nonumber\\
&=& \sum_{i=1}^{n} (n-i+1)(t_{\pi(i)} -  t_{\pi(i-1)}),
\end{eqnarray}
while respecting customers' availability profiles. 
\fi

Let $\cK \triangleq \{1, ..., m\}$ denote the set of employed robotic couriers and $c_k \in \mathbb{Z}_{+}, k \in \cK$ be their compartment capacities. Each vehicle's journey starts and ends at the central depot, however since the return trip does not influence the sought objective of minimizing the cumulative latency of deliveries, we embed $\cV$ with a dummy terminal node ${n+1}$ and write the augmented set of all vertices and edges as $\cV^+ \triangleq \{0,1, ..., n\}$ and $\cE^+ \triangleq \big\{ (i,j) : i, j \in \cV^+, i \neq j\big\}$, respectively. For each $k \in \cK$ and $(i,j) \in \cE^+$, define a \textit{binary decision variable} $x^k_{i,j}$ such that
$$x^k_{i,j} = \begin{cases}
1, & \text{if vehicle } k\text{ is directed to pass through arc } (i,j) \\
0, & \mbox{otherwise} 
\end{cases}\,.$$

With the foregoing notation, {\sc RCCVRPTW} embodies into a mixed-integer linear program of the form:

{\small \begin{align}
\displaystyle \min_{\bm{x}} \quad&   \sum_{i \in \cV} l_i  \qquad \qquad\qquad\qquad\qquad\qquad\qquad\quad(\textsc{ RCCVRPTW}~)\nonumber\\
\text{s.t.}\quad&  \sum_{k \in \cK}\sum_{j \in \cV^+} x^k_{i,j} = 1, \quad \forall i \in \cV , i \neq j\label{v1} \\
&  \sum_{j \in \cV^+} x^k_{i,j} = \sum_{j \in \cV^+} x^k_{i,j}, \quad \forall k \in \cK, i \in \cV, i \neq j \label{v2} \\
&  \sum_{j \in \cV} x^k_{0,j} \leq 1, \quad \forall k \in \cK\label{v21} \\
&  \sum_{i \in \cV} x^k_{i,n+1} \leq 1, \quad \forall k \in \cK\label{v212} \\
&  \sum_{k \in \cK}\sum_{j \in \cV^+} x^k_{n+1,j} = 0, \label{v6} \\
&  \sum_{k \in \cK}\sum_{i \in \cV^+} x^k_{i,0} = 0, \label{v7} \\
&  \sum_{i \in \cV^+}\sum_{j \in \cV^+ : j \neq i} x^k_{i,j} \leq c_k, \quad \forall k \in \cK\label{v8} \\
&  l_i - l_j + Mx^k_{i,j} \leq M - d_{i,j}, \quad \forall k \in \cK,~ i \neq j \in \cV\label{v3} \\
&  l_0 - l_j + Mx^k_{0,j} \leq M - d_{0,j}, \quad \forall k \in \cK,~ j \in \cV \label{v31} \\
&  l_i - l_{n+1} + Mx^k_{i,n+1} \leq M - d_{i,n+1}, \forall k \in \cK,i \in \cV\label{v4} \\
&  l_0 = d_{i,n+1} = 0, \quad \forall i \in \cV \label{v5} \\
&  l_{n+1} \geq 0, \label{v53} \\
&  s _i \leq l_i \leq f_i, \quad \forall i \in \cV \label{v51} \\
&  x^k_{i,j} \in \{0,1\}, \quad \forall i,j \in \cV^{+},~ k \in \cK \label{v81}\\
&  d_{i,j} \in \cU \quad \forall (i,j) \in \cE \,, \label{v9}
\end{align}}%
where $M$ stands for a sufficiently large positive constant. Given the fleet of available robotic couriers, {\sc RCCVRPTW} is targeted at finding the corresponding set of optimal delivery routes that minimize the aggregate waiting time of customers. In the above formulation, which was extended from\cite{van2000operational}, Constrs.~\raf{v1}-\raf{v7} resemble a multi-commodity network flow; Constr.~\raf{v1} ensures all the customers are serviced exactly once, Constr.~\raf{v2} guarantees that a vehicle arriving at a node will also leave that node, Constrs.~\raf{v21} to ~\raf{v7} assert that a robotic vehicle's trip originates from the depot and ends at the dummy terminal node, and each vehicle can be scheduled for at most one trip. The robotic carriers' capacity limits in terms of the number of packages are expressed through the constraint in~\raf{v8}. Constrs.~\raf{v3} to~\raf{v31} model the compatibility requirements between routes and
schedules, acting as subtour elimination constraints. Lastly, the bilateral inequality in~\raf{v51} enacts customers' preferred time windows and Constrs.~\raf{v4} and~\raf{v5} cater for the added dummy variables.

When the robotic fleet is reduced to a single uncapacitated vehicle and edge traversal times are deterministic, {\sc RCCVRPTW} simplifies to the Traveling Repairman Problem with Time Windows (TRPTW)\cite{net.3230220305}. As a special case of the TSP, the TRP (i.e. a TRPTW with time windows dropped) falls within the class of NP-hard problems and has been studied in the literature under various different names, including Minimum Latency Problem\cite{mlp}, Traveling Repairman Problem\cite{tsitsik}, Delivery Man Problem\cite{dmp} and Cumulative Traveling Salesman Problem\cite{ctsp}. Unlike TSP and VRP, where the objective is to minimize the total time required to visit all customers, TRP and CCVRP seek to minimize the cumulative waiting time of customers (i.e., service latency). In a sense, the latter problems take a \textit{customer-centric perspective} rather than a cost-oriented view as in TSP, hence the choice of cumulative latency as an objective function in {\sc RCCVRPTW}.

\if
As a special case of the TSP, the TRP belongs to the class of NP-hard problems, whereby the cumulative nature of the TRP’s objective function makes it even harder to solve compared to its TSP coun- terpart. TRP’s

The minimum latency problem (MLP) or travelling
repairman problem (TRP) is one of the most famous
customer-centric routing problems. It consists in finding a tour starting from a depot node, which minimizes the sum of the elapsed times (or latencies) to
reach a given set of nodes. The problem arises in situations in which the arrival time has a crucial role in the
customers satisfaction and it has recently attracted the
attention of the researchers, due to its importance in
applicative fields such as emergency logistics (Bruni
et al., 2018b), delivery logistics (Bruni et al., ), and
manufacturing contexts such as machine scheduling
(Bruni et al., 2019).

In such a dynamic setting, the wait for a delivery (service) may be a more important factor
than the travel cost. 
The model
is motivated by applications in which the objective is minimizing the wait for service
in a stochastic and dynamically changing environment. This is a departure from traditional vehicle routing problems which seek to minimize total travel time in a static,
deterministic environment. Potential areas of application include repair, inventory,
emergency service and scheduling models

However, solving the VRP is far from a simple task since the
problem is NP-hard [1], which implies that no algorithm capable
of finding optimal solutions in polynomial time is known.

The minimum latency problem has been well-studied in operations research, where it is also known as the traveling repairman problem and the delivery man problem. Unlike the traveling salesman problem, where the objective is minimizing maximum arrival time and therefore is server oriented, the MLP is client oriented.

The MLP is a variant of the well-known Traveling Salesman Problem (TSP) and it is known in the literature under various other names: Traveling Repairman Problem (Tsitsiklis, 1992), Delivery Man Problem (Fischetti et al., 1993), Cumulative Traveling Salesman Problem (Bianco et al., 1993) and School Bus Driver Problem (Chaudhuri et al., 2003).

Real-life applications of the MLP often arise from distribution
systems, wheresome quality criterion regarding the customer satis- faction must be focused. The MLP considers waiting times (latency) of a service system from the customer’s point of view, i.e., while in the MLP the objective is to minimize the average waiting time of each customer, in the TSP the objective is to minimize the total time required to visit all customers. In view of this, one can say that the MLP is customer oriented, while the TSP is server oriented (Archer & Williamson, 2003). Therefore, the MLP can be employed in the modeling of different types of service systems. Important practical applications can be found in home delivery services (Méndez-Dı´az, Zabala, & Lucena, 2008), logistics for emergency relief (Campbell, Vandenbussche, & Hermann, 2008) and data retrieval in computer networks (Ezzine, Semet, & Chabchoub, 2010).

The Minimum Latency Problem (MLP) is a special case of TSP, and the difference between them is that TSP is cost-oriented while MLP is customer satisfaction-oriented. This differ- ence changes the objective function of TSP, so MLP and its extensions are new study areas for researchers.

\fi

\begin{figure*}[!h]
	\centering
	\includegraphics[trim={0cm 0.3cm 0.2cm 0cm},clip, width=\textwidth]{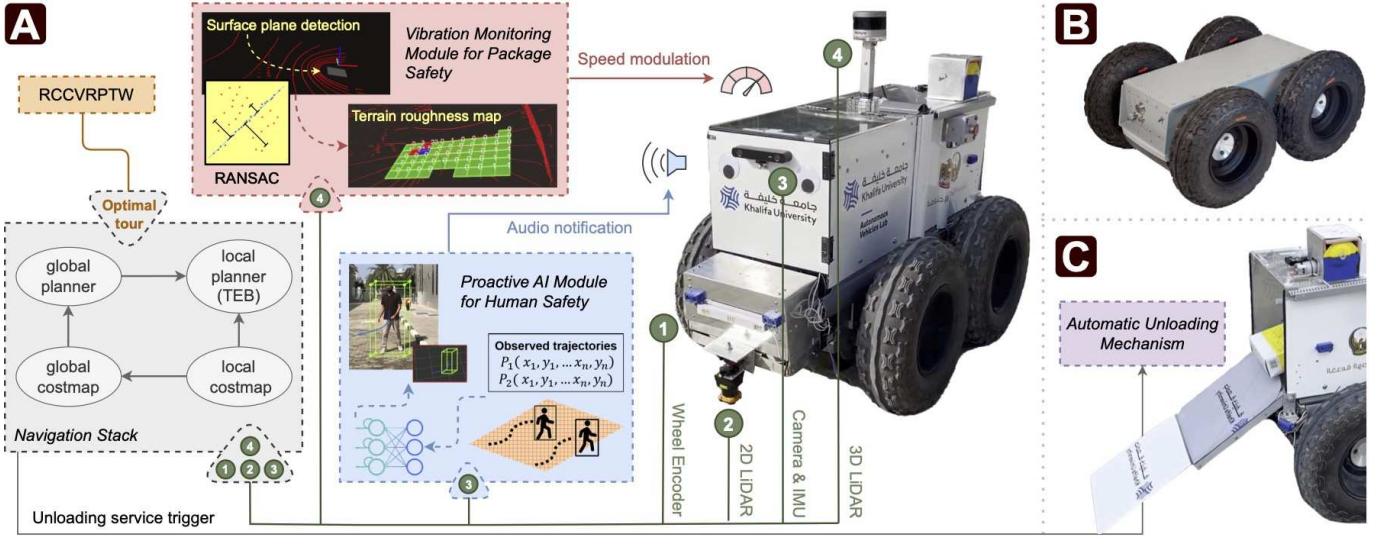}
         \caption{Infographic overview of the constructed fully autonomous LMD carrier prototype and its functionalities: A) The retrofitted vehicle and its key components; B) The bare-bone RMP 440 platform; C) Contactless multi-parcel delivery mechanism.}
	\label{fig3}
\end{figure*}

\section{Fully Autonomous LMD System}\label{sys}

This section details the developed LMD system, which appears in Fig.~\ref{fig3}, and its components, each elaborated in a separate subsection.

\subsection{System Overview}

% Explain the system without much details. list the modules, link to the corresponding sections, \ref{human}  and their intended use, what are they responsible for, explained insection this

The proposed autonomous LMD system comprises four main components: (1) the optimal logistic tour planner that computes the minimum latency dispatch routes of robotic vehicles (Sec.~\ref{prob}); (2) the vibration monitoring module (VMM) which dynamically modulates vehicles' speed based on the roughness of the driving surface to reduce vibrations (Sec.~\ref{VMMPS}); (3) the autonomous navigation stack responsible for obstacle avoidance as well as for constructing the global and local paths (Sec.~\ref{ANPSS}); (4) the proactive AI pipeline which caters for an additional layer of human safety and aids reducing deviations and interruptions in vehicles' set course of travel (Sec.~\ref{IAIPHS}). 

\subsection{Robotic LMD Courier}\label{Robotic Delivery Vehicle}

%The featured robotic delivery vehicle is designed on the philosophy of modularity that makes the hardware upgrades effortless. 
The featured robotic vehicle was assembled by retrofitting the commercial mobile platform Segway RMP 440. The platform is integrated with four lithium phosphate battery modules that maintain the 12-hour endurance of the vehicle. The maximum speed of the robot with a full payload of 100 kg is 29 km/hr. The material used to fabricate the enclosure compound is Poly (methylmethacrylate) (a.k.a Acrylic) and the aluminum L-angles are reinforced with Acrylic to provide further rigidity to the structure. At the front of the vehicle, an external ancillary battery bank enclosed with a tilt lift cover mechanism with hot-swap capability is installed to power the added equipment. As detailed below, the enclosure volume is divided into a computer and electronics zone, and package and conveyor zone. 
\begin{enumerate}
\item Computer and electronics zone: The total power requirement of the system is 485 W under normal conditions, and 650 W at peak. This zone accommodates the computing unit, sensors, and the inverter module. The employed computer requires 24 V, 20 A to operate. The sensors are safeguarded by several relays and in-line fuses. Step-up/step-down transformers are utilized to regulate the voltage and current for all the sensors, motors, micro-controller, and motor controllers. A power inverter outputting a 3 KW pure sine waveform is installed to support accessories that run on alternating currents.
\item Package and conveyor zone: The unloading mechanism relies on a conveyor belt driven by a DC motor which can support up to 8 kg of load. Three packages of dimensions 200mm x 200mm x 200mm can be fit onto the belt. The contactless delivery is achieved via a two-fold ramp mechanism that extends the ramp length twice. The ramp is unfolded at an angle, and the package slides over to the ground. The delivery process of a single parcel takes less than 20 seconds. The choice of the current unloading setup owes primarily to its simplicity and cost, whereas for future deployments the mechanism can be substituted by a robotic arm-based retrieving mechanism that can detect and place the intended package on the ramp.

%The proposed delivery mechanism can easily be extended by providing an anti-theft mechanism for the remaining packages and 

\end{enumerate}

%As illustrated in Fig.~\ref{fig3}, the vehicle relies on 2 LIDars, a Zed camera and NUOVO ipc. Justify IPC. The top 3D, highl;ight that we can replace it with 2d. The 2D lidar, is utilised mainly for obstacle and curb detection, justifies the positining. For the effective use of the VMM module, we avail of an additional 3D lidar placed next to 2D lidar.  \\

As Fig.~\ref{fig3} depicts, the vehicle is equipped with a 2D LiDAR (Hokuyo UTM-30LX), 3D LiDAR (Velodyne Puck Lite), stereo camera (Stereolabs ZED 2), and an IPC (Nuvo 8108GC-XL Industrial PC). The IPC can operate at high ranges of ambient temperature and humidity as well as withstand extensive shock and vibrations, which duly suits the current deployment environment in Abu Dhabi. The 3D LiDAR, which is part of the vehicle localization system, was utilized due mainly to its long detection range ($100$ m) and could have been replaced with a 2D alternative with a comparable range since only a single laser scan layer was processed. The positioning of the bottom LiDAR was set in order to capture curbs and otherwise overlooked low obstacles, effectively allowing to avoid computationally intensive cloud/laser filtering or surface segmentation techniques. For an effective implementation of the VM module (Sec.~\ref{VMMPS}), we avail an additional 3D LiDAR placed at the front of the vehicle at an inclination of 15 degrees with respect to the ground, which allows capturing the immediate in-front driving surface. The Zed 2 camera, which supports spatial object detection and features an in-built IMU, was employed to complement the autonomous navigation stack and the proactive AI module for human safety (Sec.~\ref{IAIPHS}).

%a 2D LiDAR 

%since our 2D localization method proved desirable results, its  were not made use of and the LiDAR itself can be switched with .
%commonly employed in AV applications because of its ruggedness and aptness for usage in a high ranges of operating temperatures, humidity, shock, and vibrations, .  The positioning of the Hokuyo 2D LiDAR was expressly allocated to be below the height of the curbs on the campus of 14 cms, this allows us to forego any obligation to implement a more complicated detection method of the sidewalk, such as segmentation or cloud/laser filtering  of the sidewalk and the drivable ground. In addition, the 2D LiDAR acts as an obstacle detector with a range of around 6 meters as specified in the local cost map, and approximately 180 degrees field of view. 

\subsection{Autonomous Navigation and Platform Safety Stack}\label{ANPSS}

The assembled navigation stack operates based on the time-series sensor readings received from the LiDARs, IMU, and wheel encoders as illustrated in Fig. 3. Prior to deployment, the operational area of the system was mapped to generate the global costmap, which is a two-dimensional representation of static obstacles in the environment. The global planner takes as an input a set of way-points produced by RCCVRPTW (i.e., the optimal logistic sequence) and processes the set iteratively, activating the automatic unloading mechanism whenever the corresponding navigation goals are reached. For local path planning, we resort to the Time-Elastic-Band (TEB) algorithm; a decision taken under extensive experimentation with existing planners in ROS, such as the Dynamic Window Approach (DWA) and the base local planner. TEB relies on the local costmap generated from real-time LiDAR readings to design locally optimal paths allowing for collision-free navigation among detected static and dynamic obstacles. However, the default parameters of the local and global planners in ROS may require judicious fine-tuning to satisfy the desired risk tolerance level and use-case specifics. Since in the current LMD system platform safety is prioritised, we adjust the parameters in a conservative manner as detailed in Sec.~\ref{sste}.

%The roughness estimate outputted by the VMM is used to modulate robotic couriers' speed in real time, while the proactive AI module for human safety activates the speakers to play an audible message for early notification of pedestrians.

\subsection{Proactive AI Pipeline for Human Safety}\label{IAIPHS}

To reinforce the human safety component within the proposed LMD system, we develop an AI-boosted proactive module to predict pedestrians' trajectories and minimize the risk of collisions. Pedestrian trajectory prediction is an extensively researched topic that has been studied for various applications, including pedestrian safety, human-tracking, and crowd surveillance. The challenge here lies primarily in the complexity of human motion, which conventionally was modeled via physics-based approaches with handcrafted techniques ~\cite{PhysRevE.51.4282, 10.1016/j.patcog.2011.04.022, 10.1109/CVPR.2011.5995468}. Although successful to an extent, such methods are limited in their representation and scalability \cite{Nikhil2018ConvolutionalNN}. Most recent trajectory prediction methods resort to deep learning (DL) techniques, which allow to alleviate some of the challenges faced with handcrafted methods. Recurrent Neural Network (RNN) is one of the prevalent DL methods in the literature on trajectory prediction. RNNs are commonly used to predict sequential data such as trajectory. Its more advanced variant, known as the Long Short-term Memory (LSTM) network, have gained traction as an efficient method for pedestrian trajectory prediction 
in complex, dynamic scenes\cite{gupta2018social, Alahi_2016_CVPR, BiTraP}.
\begin{figure}[!b]
	\centering
    \includegraphics[width=\columnwidth]{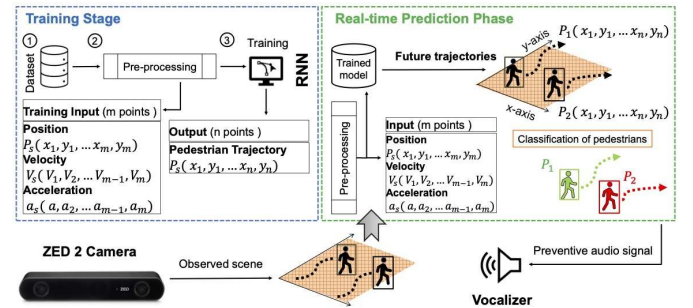}
	\caption{A summary of the proposed proactive human safety pipeline with trajectory prediction and vocalizer.}
	\label{humansafety}
\end{figure}

For the purposes of the current study, we employ the model proposed in \cite{BiTraP} referred to as Bi-directional Pedestrian Trajectory Prediction with Multi-modal Goal Estimation Model (BiTrap). BiTrap is a goal-oriented conditional variational autoencoder hinging on an RNN-based trajectory prediction model. Thanks to the inbuilt advanced bi-directional decoder, BiTrap allows for long-term trajectory prediction. As an input, the model takes pedestrians' observed trajectories as well as motion data such as velocities and accelerations.

In the proposed module, the ZED 2 camera depicted in Fig.~\ref{humansafety} was employed to detect, locate, and track pedestrians. Particularly, bounding boxes of all pedestrians in the field of view of the camera are extracted. These boxes contain the 3D coordinates of the pedestrians in reference to the camera. The tracked information is saved as observed trajectories, which are then fed to BiTrap. Following the related works in~\cite{gupta2018social, Alahi_2016_CVPR, BiTraP,MTN,PIE}, $8$ data points from a pedestrian's observed trajectory are taken as input, and $12$ data points are generated as output. Depending on the predicted trajectories of pedestrians and their observed distances (from the vehicle) a preventive audio signal is produced via a ROS service (dubbed vocalizer). Here, the audio message serves as an early notification to pedestrians, intending to circumvent potential collisions with the vehicle as well as minimize interruptions/deviations in the vehicle's set course of navigation.

For the sake of analysis, we classify pedestrians into three categories: red, blue, and green. The red category highlights pedestrians that are in a dangerously close-proximity to the vehicle as determined by a preset threshold. Pedestrians in blue are those whose future trajectory predicted by BiTrap model collides with the vehicle's planned path (unless the speed or trajectory of the pedestrian changes). The green category is for pedestrians who are at a safe distance from the vehicle and no collision is foreseen. The vocalizer is set to react for blue and red classes.  

% Here, a pedestrian is defined to be in close proximity to the vehicle based on a  preset threshold specified by the system user ({currently 0.5 meters}). If a pedestrian is within a close proximity and their predicted future trajectory is towards the vehicle, the Vocalizer is activated.
 
% For a comprehensive analysis of collision-risk, we  classify pedestrians into 3 categories, i.e red, blue, green. The first class is the red or danger class,in which a pedestrian is in dangerously close-proximity to the vehicle as determined by the threshold. Pedestrians in the blue class are those whose future trajectory predicted from BiTrap model collides with the future trajectory of the vehicle. This means there is a possibility of collision between the pedestrian and vehicle unless one of the two changes speed or trajectory. The green class is for pedestrians who are in safe distance from the vehicle and no collision is predicted between their short-term future path and vehicle's path. 
% The Vocalizer is set to react for blue and red cases.  

\subsection{Vibration Monitoring Module for Package Safety}\label{VMMPS}

% One aspect that can affect the quality of a vehicle's journey is the quality of the road it is driven on
%The quality of the drivable surface plays an important factor on the vehicle's journey, 
% Driving on rough terrains can induce vibrations on robotic vehicles, which depending on the severity, can cause damages to the parcels carried or the vehicle's hardware itself. Suspension systems can dampen the intensity of these vibrations, but only to a limited extent. They also require dedicated hardware to be installed, which incurs additional cost and complexity, and may not be feasible in some situations. In general, the vibrations sustained by a vehicle correlate with its speed: the higher the speed, the higher the intensity. An approach which can tackle both platform and package safety would be to modulate the speed based on the perceived roughness on the surface. Accordingly, the surface roughness should be adequately quantified.
When dispatched on uneven terrains, robotic vehicles may endure vibrations that could result in damage to the vehicle's hardware or the packages carried. While suspension systems can alleviate these concerns, they require dedicated hardware, which can add complexity and cost. Furthermore, installing such systems may not be possible in some situations, or might not be worth the investment if the vehicle generally drives on an even terrain and only occasionally encounters rough patches or segments. Generally, the severity of the vibrations that a robotic vehicle sustains is directly related to its speed. Therefore an effective way to improve both the safety of the vehicle and the packages it carries is to modulate its speed based on the roughness of the driving surface. 
% However, to achieve this, it's essential to have an accurate way to quantify the roughness of the surface.

% these is a limitation on the extent of

Some relevant research work in this area concentrates on road cracks and pothole detection. Anand and Gupta \cite{anand2018crackpot} presented an end-to-end DL approach to detect road cracks by texture and spatial features on the image. The approach employs Seg-Net \cite{kendall2016bayesian, badrinarayanan2015segnet} for semantic segmentation to extract road pixels, followed by a neural network to detect cracks. On the other hand, Aggarwal and Jain \cite{9002060} resort to basic image processing algorithms to capture cracks and potholes.

\begin{figure}[!t]
	\centering
	\includegraphics[trim={0cm 0.2cm 0cm 0cm},clip, width=\columnwidth]{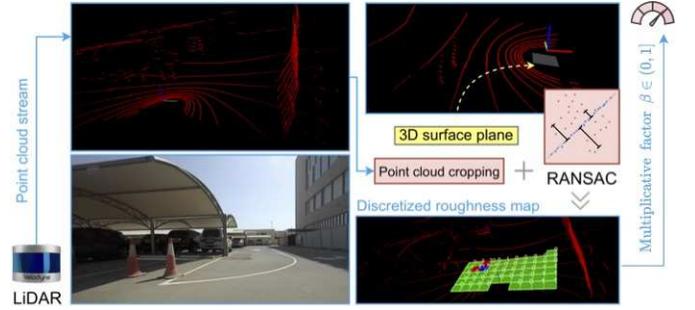}
	% \caption{{\color{red}[M]} Visualization of a 3D roughness estimation map for a real-time experiment has been done by a delivery robot using a stereo camera and LiDAR sensors. }
 	\caption{Illustration of the proposed Vibration Monitoring module with speed modulation. Point cloud data from a 3D LiDAR is parsed through RANSAC to detect the surface plain, based on which the discretized roughness map is constructed in real time.}

	\label{fig:roughness}
\end{figure}

Detecting cracks and potholes on a driving surface provides only a partial understanding of its condition since it doesn't reflect the overall smoothness of the ground. To address this limitation, we leverage a LiDAR to generate a 3D grid map that captures the surface roughness more accurately. Initially, the LiDAR's point cloud is cropped to cover the area in front of the vehicle. Then, the RANSAC algorithm \cite{fujiwara2013evaluation} is invoked to obtain a 3D plane of the driving surface. The algorithm iteratively generates various plane parameters and selects the best one fitting the highest portion of the given point cloud. We then quantify the surface roughness by measuring the perpendicular Euclidean distance between each impact point on the surface and the fitted plane.
Since the LiDAR's point cloud resolution is not sufficiently high, we split the surface into smaller boxes ($1$m$ \times 1$m) and compute the average roughness in that area. The calculated roughness score can be utilised in two ways: (1) It can be embedded into the navigation stack as a local cost map, allowing the local planner to generate a path that passes through the lowest cost boxes while adjusting speed if necessary; (2) Alternatively, only the boxes directly in front of the wheels' orientation can be considered and the vehicle's speed can be adjusted based on the roughness of those boxes. Overall, this module creates a trade-off between speed and ride quality that can be fine-tuned depending on the application specifics.

In the proposed method, overviewed in Fig. \ref{fig:roughness}, we sample the boxes two meters directly in front of the robotic vehicle. The boxes with roughness scores from $0$ to $5$ cm are highlighted in green as an indication of a smooth surface. The boxes highlighted in blue represent a moderate roughness with a score between $5$ to $20$ cm, whereas rough surfaces with a score higher than $20$ cm are colored in red. These thresholds were determined based on real-world experiments on various driving surfaces and can be adjusted based on the adopted platform and application. The VMM throttles the vehicle's speed by\textit{ a multiplicative factor} $\beta \in (0,1]$ according to the computed average score of these regions. For high roughness scores, we set $\beta = 0.5$, while for moderate and low roughness $\beta = 0.75$ and $\beta = 1$ were considered, respectively.

\begin{figure*}[!h]
	\centering
	\includegraphics[width=\textwidth]{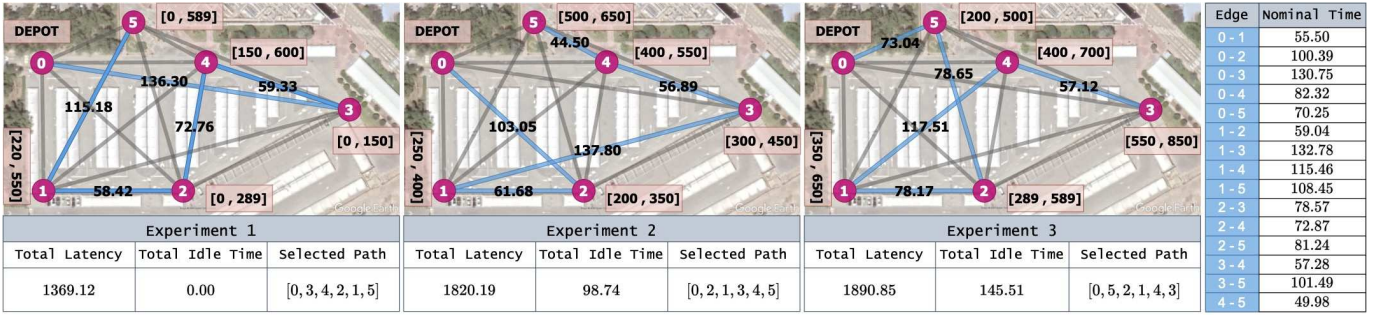}
	\caption{Visualized results of the performed real-world experiments. The edges highlighted in blue represent the optimal path selected by {\sc RCCVRPTW}. The numbers on the edges encode the actual travel time (in seconds) of the robotic vehicle during the experiments, while the table on the right lists the estimated nominal times.}
	\label{frm2}
\end{figure*}
% \usepackage{graphicx}

\section{Field Tests and Simulation Studies}\label{sec:exp}

To test the effectiveness and practicality of the proposed autonomous LMD system, a series of real-world trial experiments were conducted in which the robotic courier presented in Sec.~\ref{sys} was deployed to deliver packages to five customers autonomously. Video footage of some of the test runs is available online\footnote{\url{https://youtu.be/-9Zams67hNY}}. Complementary to these real-world tests, in Sec.~\ref{scalb} simulation studies are provided to examine the scalability of {\sc RCCVRPTW}'s MILP model established in Sec.~\ref{prob} with respect to the number of vehicles and customers.

\subsection{Setup and Scenarios of Trial Experiments}\label{sste}

The first set of experiments were performed in Khalifa University's staff parking lot. The area is around $10,000$ square meters, and contains shaded parking slots that are in clusters around the middle. The parking lot is paved and routinely maintained and features mixed traffic of vehicles and pedestrians. The area mapping was performed with a Simultaneous Localization and Mapping (SLAM) package that is based on OpenSlam's Gmapping particle filter. Gmapping provides a laser-based 2D SLAM solution to produce a 2D occupancy grid map from laser and pose data collected by the vehicle, the vehicle was driven around the perimeter and in between the parking lot sections to generate this occupancy grid. To facilitate the global path planning within the autonomous navigation stack, the global occupancy grid was manually edited to highlight the parking areas, sidewalks around the perimeter of each parking cluster, as well as sidewalks around the perimeter of the whole lot; these low features were not properly captured by the 3D LiDAR, which was placed in a high vantage point in order to produce a reliable occupancy grid for localization. If the occupancy grid was used without editing, the global planner would produce paths that often would overlap with untraversable areas. Although the local planner can adjust the path dynamically on-the-fly to bypass newly detected unmapped obstacles, the resulting trajectories would likely be jagged and more lengthy. For enhanced safety of robotic carriers, the autonomous navigation stack parameters in ROS were tuned as follows. The buffer (inflation) zone of obstacles was set to 1 meter, the maximum speed was set to 1.2 m/s, and the local costmap dimension was set to 6x6 meters.

%To assure platform safety, we adapted the autonomous navigation stack parameters in ROS as follows. The buffer (inflation) zone between obstacles was set to $1$ meter, the maximum speed was set to $1.2$ m/s and local costmap dimension was set to 6x6 meters.

% this is because the 3D LiDAR used for mapping needs to be placed in a high vantage point in order to capture a proper occupancy grid for localization, which does not allow it to capture these low features, and without editing the occupancy grid the global planner would produce paths that often would overlap with undrivable paths. Even though the local planner can compensate for this shortcoming and adjust the path dynamically while driving, that would produce jagged trajectories and would add unneeded time costs to the traversal.

One can observe from Fig.~\ref{fig3} that the current implementation does not rely on GPS for localisation or navigation. The integration of the GPS data into the vehicle's pose proved no noticeable improvement while adding more computational and hardware requirements. This owes mainly to the characteristics of the testing area, which was feature dense and yielded sufficiently accurate localisation results with LiDAR. In addition, a GPS-less solution would prove more useful since it could be extended into areas with unreliable GPS coverage or GPS-denied environments such as indoor locations.
% It's noteworthy that a GPS could be integrated to the localization and navigation solution, either by utilizing a map-less approach with an empty global occupancy grid, or by integrating the data to improve the pose of the vehicle, however, the map-less option would require a full, more intensive reconstruction of the global occupancy grid for a smooth global path generation,

The delivery nodes were allocated in the corners of the parking lots as portrayed in Fig.~\ref{frm2}. To determine the pairwise nominal travel time between the nodes (to be inputted to {\sc RCCVRPTW}) the vehicle was autonomously driven 10 times along each edge (in both directions) and the average value was taken. As one demonstration, the proposed autonomous LMD solution is tested on three scenarios with varying delivery time preferences. In the first scenario, the opening time windows of the customers are mostly relaxed (i.e., large time ranges where customers accept deliveries), while the second and third had stricter temporal constraints. To discern between the results, the corresponding experiments are labeled \textbf{E1} to \textbf{E3}. For each experiment, the vehicle's optimal logistic route was obtained by invoking the Gurobi optimizer on the respective MILP program of {\sc RCCVRPTW}.
\begin{figure*}[!b]
	\centering
	\includegraphics[trim={0cm 0.3cm 0.2cm 0cm},clip, width=\textwidth]{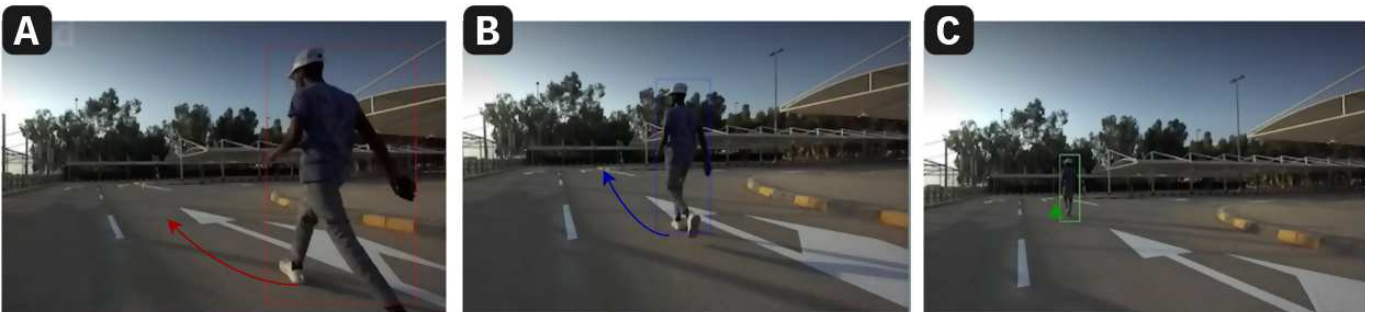}
         \caption{Snippets of a pedestrian interaction with the vehicle during \textbf{E3}: A) The pedestrian moves dangerously close to vehicle. B) The pedestrian moves towards the vehicle's path C) The pedestrian moves away from the vehicle.}
	\label{Pedestrian movements}
\end{figure*}
The site in which the first set of experiments were conducted is well paved and routinely maintained, thus in order to investigate the proposed VMM's performance two additional experiments, labeled \textbf{E4} and \textbf{E5}, were conducted (which are summarized in Fig.~\ref{vibrations}). These new experiments were performed in a different area that featured an unpaved and uneven sandy terrain. The LMD vehicle was autonomously driven in a $50$ meters straight path. The VMM's speed modulating parameter was set to $\beta = 0.75$ when the ground's estimated roughness value falls in-between $5$ and $20$ cm.

% Justify why in a different area .....
%Each edge is considered bi-directional and this was considered while collecting the data by traversing in both direction of the edge. 
%To demonstrate the proposed autonomous LD solution, three scenarios were considered with varying delivery time preferences. .

\subsection{Results}
\subsubsection{LMD Performance} 

As evidenced by Fig.~\ref{frm2}, the customers' delivery schedules (i.e., time windows) were respected in all the experiments. There was no idle time in \textbf{E1}, where \textbf{E2} and \textbf{E3} had $98.74$ and $145.51$ seconds of idle time, respectively. It is to be noted that idle time is not incorporated into the objective function of the current formulation. For future investigations, a bi-objective optimization problem can be considered to additionally minimize idle time. The actual traversal time in \textbf{E1} to \textbf{E3} differs from the estimated nominal time, which is due to stochastic factors such as pedestrian movements and sensor noise. This substantiates the need to account for uncertainties in the formulation. In the experiments performed, the observed deviations fell within the range ($\pm 5$ seconds) as evidenced by Fig.~\ref{frm2}.

% 1. no violation of time windows
% 2. idle time, hint that we can also optimize idle time, or use idle time to rechage the vehicle
%  as indicated by Fig.\ref{frm2}, the 

%  as indicated by fig.\ref{frm2}, customers time windows were respected. notice that in experimetn 1, the vehicle wasn't idle. we note that the idle time can be incorporated in the objective function. Observe that, the actual time differs from the estimated nominal, argue why, pedesterians, navigation stack. this substantiates the need to account for uncertainities in the formulation. While in the fiurst two experiments, the deviations fall within the anticipated range (+- 5secs.), in the third experiment the travel time from node 1 to 2 far exceeded the nominal time. argue why ? 

\subsubsection{Package Safety}
\begin{figure}[!h]
	\centering
	\includegraphics[trim={0.22cm 0.2cm 0.29cm 0cm},clip, width=\columnwidth]{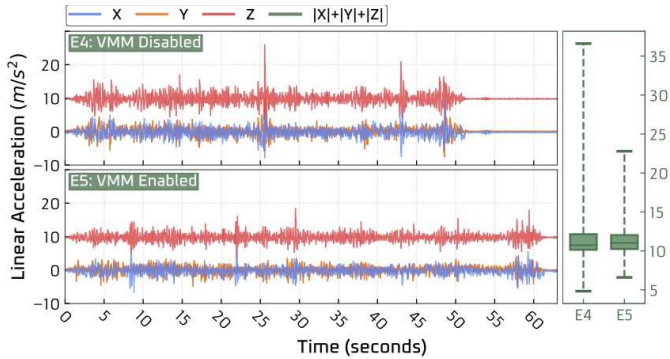}
	\caption{Comparison of the recorded vibrations in \textbf{E4} (VMM disabled) and \textbf{E5} (VMM enabled). The right-most subplot displays the range of vibration values in each experiment.}
	\label{vibrations}
\end{figure}

The results of \textbf{E4} and \textbf{E5} are highlighted in Fig.\ref{vibrations}, the plot shows the linear acceleration readings from the LMD vehicle's IMU in the X, Y, and Z axes. \textbf{E4} shows the readings with the VMM disabled, while \textbf{E5} displays the readings with the VMM enabled. It is observed that with the VMM disabled, there are more spikes in the IMU's readings, and the intensity of the vibrations is higher than compared with the VMM being enabled. Since there is no speed modulation in \textbf{E4}, the vehicle took 50 seconds to traverse the 50-meter straight path autonomously, compared with the 62 seconds in \textbf{E5} where the VMM was frequently engaging to throttle the speed based on the roughness estimates computed by the perception system. This yielded around $40\%$ decrease in the magnitude of vibrations.

% explain why e4 is shorter.
% Vibrations 1.5 times higher than in the case with VMM enabled
% explain 
%mention that edpending on the application the speed modulation factor can be adjusted, e.g. for highly fragile parcels
%note that instead of relying solely on speed modulation one can design an embedded VMM module. Specifically, the modeule can be integrated with a local planner such that the local path is adjusted based on the roughness map besides speed modulation. 

\subsubsection{Real-time Intent Prediction and Early Notification of Pedestrians}
 In this section, we provide a qualitative and quantitative analysis of the developed proactive AI Pipeline for human safety based on the results of \textbf{E3}, video footage of which is available online (\url{https://youtu.be/-9Zams67hNY}). The experiment was performed in a relatively less crowded parking lot, however, various interaction scenarios were captured.

 As mentioned in Sec.~\ref{IAIPHS}, we discern pedestrian bounding boxes into three color-coded classes: green, blue, and red. The green box indicates a pedestrian being at a safe distance with no collision predicted in the near future. The blue class indicates a pedestrian being at a safe distance from the vehicle, but their predicted trajectory and the near future trajectory of the vehicle might collide. Lastly, the red bounding box signifies an imminent danger of collision with the vehicle, due mainly to the proximity of the pedestrian to the vehicle. 

% For evaluation, we consider  {\color{red} experiment 3} results as provided in the video. In this experiment, 
% During the experiment, the vehicle had five total instances of interactions with pedestrians. Three instances on single pedestrians, one instance with two pedestrians, and one instance with a group of pedestrians. 
In \textbf{E3}, the vehicle had five total instances of interactions with pedestrians: three with single pedestrians, one with two pedestrians, and one with a group. Since trajectory prediction is a continuous process, we evaluate the developed predictive model by dividing each interaction into three distinct phases. The first phase assigns a class label to the pedestrian after initial detection, using the aforementioned green, blue, and red bounding boxes. The second phase assigns a label to the pedestrian halfway through the interaction, based on their predicted behavior and trajectory. The third and final phase assigns a label just before the pedestrian leaves the view or vicinity of the vehicle. Interactions refer to the time from pedestrian detection to the moment when they are no longer within 20 meters of the vehicle.
% Since prediction is a continuous process, it is hard to evaluate the model in real-time. Therefore we divide each interaction with a pedestrian into three phases. The first phase prediction refers to the class label given to a pedestrian after first detection. The second phase refers to the label given to a pedestrian halfway through the entire interaction time. Here, interaction refers to time between the detection of the pedestrian to the time the pedestrian is no longer in the vicinity of the vehicle. In our case, the detection limit is set to be 20 meters. The third and final phase is the class label given to a pedestrian just before leaving the view or vicinity of the vehicle.

\begin{table}[!b]
\centering
\resizebox{\columnwidth}{!}{
\begin{tabular}{|c|c|c|c|c|}
\hline
\textbf{Interaction} & \textbf{Number of pedestrians} & \textbf{Detected Pedestrians} & \textbf{Class Label} & \textbf{Expected Label} \\ \hline
1                    & 1                              & 1                             & G-G-G                & G-G-G                   \\ \hline
2                    & 1                              & 1                             & G-B-R                & G-B-R                   \\ \hline
3                    & 3+                             & 2                             & G-G-G                & G-G-G                   \\ \hline
4                    & 1                              & 1                             & R-B-G                & R-B-G                   \\ \hline
5                    & 2                              & 2                             & G-B-G                & G-B-G                   \\ \hline
\end{tabular}}
\caption{Performance of the adopted BiTrap model for inferring pedestrians' future trajectories. The three labels (G - Green, B - Blue, and R - Red) delimited by dashes (e.g., G-B-G) represent the class labels in the three phases of interaction.}
\label{tab: interactive_module}

\end{table}

The results of the experiment are listed in Table~\ref{tab: interactive_module}. In the first interaction, a pedestrian initially positioned around 15 meters from the vehicle moved further away. As anticipated, the module labels the pedestrian green throughout the interaction interval. In the second interaction, a pedestrian initially moves towards the vehicle and then stops on the vehicle's path while it continues approaching. The proactive module labels the pedestrian as safe at first since the pedestrian was far away and hinted signs of moving left. When the pedestrian was standing still the output label started alternating between green and blue. This is because the observed trajectory for a standing pedestrian does not provide sufficient context on their future intention. After a while, however, the pedestrian's position became too close to the vehicle and therefore the red label was assigned. In the third interaction, a group of pedestrians were standing on the left side of the vehicle in a shaded parking. Due to poor visibility, only two of the three were detected. Since all of them were standing in a safe distance, they were labeled green.

% In the fourth interaction, a pedestrian comes from the right side of the vehicle and moves straight. At first, the pedestrian was in close proximity to the vehicle (less than 0.5 meters) and correctly labeled to the red class. After a few seconds the pedestrians moves forward and away from the vehicle, but their path still intersects the vehicle's planned path. This means there might be possible collision between the two in near future. Therefore a blue label is assigned to the pedestrian. Later, the pedestrian gains more distance and gets further from the vehicle, which changes the label to green. Snippets of the interaction are shown in Figure~\ref{Pedestrian movements}.

In the fourth interaction, illustrated in Fig.~\ref{Pedestrian movements}, a pedestrian approached the vehicle from the right side and was labeled red due to their close proximity (less than 0.5 meters). As the pedestrian moved away, the label turned blue since their path still intersected with the vehicle's planned trajectory. Eventually, the pedestrian moved further away, resulting in a change of their classification to green. During the final interaction, two pedestrians emerged on the right side of the vehicle and were labeled as green. As they approached the vehicle and seemed to be crossing its path, their labels changed to blue. However, they eventually came to a stop without crossing the vehicle's path, and the label changed to green.

% In the final interaction two pedestrians intending to cross through the path of the vehicle while close to the vehicle are detected. When the pedestrians get close to the vehicle they stop. Since the pedestrians are further away at the time of detection, they are labeled green, which changes blue as the get closer. After learning the pedestrians are not moving and are not dangerously close, the vehicle changes their label to blue and continues it's path.

\subsection{Scalability Analysis}\label{scalb}

\begin{figure}[!b]
	\centering
	\includegraphics[trim={0.22cm 0.2cm 0.29cm 0cm},clip, width=0.99\columnwidth]{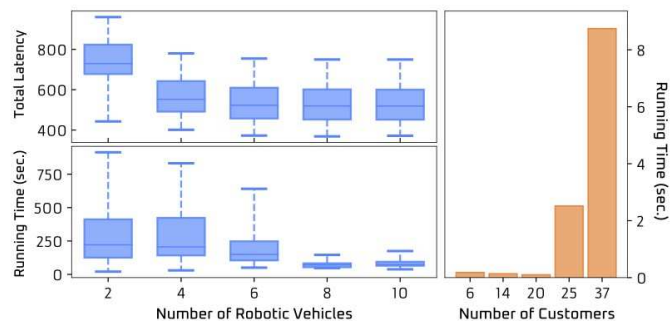}
	\caption{Performance of the Gurobi solver on higher-dimensional instances of {\sc RCCVRPTW}. The subplots on the left depict the recorded running time and objective value against the number of robotic vehicles. The rightmost subplot reports the runtime against the number of nodes.}
	\label{lastlastdecember}
\end{figure}

In view of practical limitations, {\sc RCCVRPTW} instances considered in the above-reported field tests were confined to a single vehicle and $5$ nodes (customers). To shed light on the empirical computational cost of {\sc RCCVRPTW} for higher-dimensional inputs, this section analyzes the results of numerical simulations (summarized in Fig.~\ref{lastlastdecember}) with up to 37 customers and 10 robotic LMD couriers. The simulations were coded in Python and carried out on an Intel i9-9900k CPU.

We evaluate the presented MILP model's performance through a series of simulations based on two different sets of instances. In the first case study, the results of which appear on the leftmost two subplots in Fig.~\ref{lastlastdecember}, {\sc RCCVRPTW}'s output is analyzed on a set of $50$ randomly generated synthetic instances with $n=11$ customers and vehicle capacities set to $c_k=\ceil{2\cdot\frac{n}{|\cK|}},~\forall k \in \cK$, considering $|\cK|$ being varied from $2$ to $10$ in steps of $2$. The second case study, visualised in the rightmost subplot of Fig.~\ref{lastlastdecember}, investigates the model's scalability with respect to the number of nodes on $5$ instances\footnote{$\big\{\text{RC\_204.3, RC\_207.4, RC\_205.1, RC\_203.1, RC\_202.1}\big\}$}, generated by Potvin and Bengio \cite{doi:10.1287/ijoc.8.2.165} from Solomon’s RC2 VRPTW benchmark set, where $|\cK|$ was set to $6$ and $c_k = 10, ~\forall k \in \cK$.

As Fig.~\ref{lastlastdecember} indicates, when applied to {\sc RCCVRPTW} the solver's execution time does not scale linearly with the problem size and may vary significantly depending on the instance data. While for the RC2 instances, the running time did not exceed $10$ seconds, in the case of random instances with an equal number of LMD vehicles ($|\cK|=6$) the execution time averaged approximately to $200$ seconds. On the positive side, the observed running time decreased drastically as the fleet size $|\cK|$ increased (since the instance complexity is more relaxed). We remark that for large-scale instances with more than $40$ customers, solving {\sc RCCVRPTW} optimally can be intractable. In such situations, one can resort to efficient heuristic/meta-heuristic approaches such as the Hybrid Genetic Search algorithm proposed in~\cite{VIDAL2022105643}.

\section{Conclusion}

Aiming to inform and advance the design of effective autonomous last-mile logistic systems, we developed and experimentally demonstrated a customer-centric, fully-autonomous LMD system for small urban communities. The presented system relies on self-driving robotic vehicles, which are boosted with three auxiliary modules for enhanced usability and added safety: (1) A proactive audio service for real-time intent prediction and early notification of pedestrians; (2) A reactive perception
system for vibration monitoring and parcel safety; (3) A contactless multi-package delivery mechanism. The results of proof-of-concept operation tests illustrate the effectiveness of the proposed system in providing fully-autonomous parcel delivery services while minimizing the latency of deliveries. In all the field tests conducted, the developed autonomous robotic courier navigated safely and intelligently in the presence of pedestrians, cars and dynamic obstacles. One promising extension is to integrate the proactive AI module and the reactive perception system within the navigation stack planners to allow for more optimised planning of 'local' delivery paths. Additionally, future work includes extending the proposed system to the setting with cooperative LMD robots as well as incorporating battery recharging and energy constraints.

\bibliographystyle{IEEEtran}
\bibliography{reference}

\end{document}